\documentclass{article}
\pdfoutput=1
\usepackage{microtype}
\usepackage{graphicx}
\usepackage{subfigure}
\usepackage{booktabs} % for professional tables

\usepackage{hyperref}

\usepackage[accepted]{icml2018modified}

\icmltitlerunning{A Novel Variational Autoencoder with Applications to Generative Modelling, Classification, and Ordinal Regression}

\usepackage{amssymb,amsthm,amsmath}
\usepackage{color}
\usepackage{url}
\usepackage{natbib}
\newcommand{\figureDir}{.} % 

\newcommand{\CUT}[1]{} % drafting helper, got this from Chris Williams

\newcommand{\transpose}[1]{{#1}^{\top}}
\newcommand{\condProbP}[2]{\ensuremath{\left.p\!\left({#1}\right|{#2}\right)}}

\newcommand{\condProbPP}[2]{\ensuremath{\left.p_{\theta}\!\left({#1}\right|{#2}\right)}}
\newcommand{\condProbQQ}[2]{\ensuremath{\left.q_{\phi}\!\left({#1}\right|{#2}\right)}}
\DeclareMathOperator*{\argmax}{argmax}
\begin{document}

\twocolumn[
\icmltitle{A Novel Variational Autoencoder with Applications to\\ Generative Modelling, Classification, and Ordinal Regression}




\begin{icmlauthorlist}
\icmlauthor{Joel Jaskari}{affil}
\icmlauthor{Jyri J. Kivinen}{affil}
\end{icmlauthorlist}
\icmlaffiliation{affil}{Department of Computer Science, Aalto University, Espoo, Finland}
\icmlcorrespondingauthor{Joel Jaskari}{joel.jaskari@aalto.fi}
\icmlcorrespondingauthor{Jyri Kivinen}{jyri.kivinen@aalto.fi}
\icmlkeywords{Machine Learning, Deep Learning, Variational Autoencoders, Deep Generative Models}

\vskip 0.3in
]



\printAffiliationsAndNotice{}  % leave blank if no need to mention equal contribution

\begin{abstract}
We develop a novel probabilistic generative model based on the variational autoencoder approach. Notable aspects of our architecture
are: a novel way of specifying the latent variables prior, and the introduction of an ordinality enforcing unit. We describe how to do supervised, unsupervised and semi-supervised learning, and nominal and ordinal classification, with the model. We analyze generative properties of the approach, and the classification effectiveness under nominal and ordinal classification, using two benchmark
datasets. Our results show that our model can achieve comparable results with relevant baselines in both of the classification tasks.
\end{abstract}

\section{Introduction}
We consider the problem of modeling statistical structure occuring in data using variational autoencoder~\citep{kingma+welling14,rezende+14} architectures, for generative and discriminative tasks. Such models define two kinds of directed and partially stochastic neural networks, often termed the generation (/decoder) and the recognition (/encoder) network (similar to as in the classical approach taken by Helmholtz machines~\citep{dayan+95,hinton+95}), each of which may contain multiple layers of non-linearly processing units.

While exact inference and likelihood computation is in general intractable for these models, generative synthesis is straightforward from the generation network (similar to the classical density networks~\citep{mackay+gibbs98}), and with the innovations of the variational (evidence) lower bound reparametrization~\citep{kingma+welling14,rezende+14}, the network pairs can be tuned concertedly under gradient based training, with the recognition network learning to do approximate inference, and defining a distribution aligning with the true posterior when the bound is (/becomes) tight.

Although relatively recent\CUT{(proposed in~\citep{kingma+welling14,rezende+14})}, such architectures have been highly popular within the deep learning field, with several variations being proposed for a variety of tasks. Such include extensions to stochastically deeper architectures (such as taken by~\citet{kingma+14,maaloe+16,sonderby+16}), to tighter lower bounds to optimize such as in~\citet{burda+16}, and for considering semi-supervised classification tasks, such as in~\citet{kingma+14,maaloe+16}, a task which we also consider with our proposed new approach.

Our main contributions are the following: i) we propose a novel and simplistic variational autoencoder architecture, which can be used for unsupervised, semi-supervised, and supervised learning; ii) we propose a novel kind of unit to be used for ordinal classification; iii) we analyze and demonstrate the effectiveness of our approach in comparison to similar complexity models in a variety of tasks: generative modelling, nominal classification, and ordinal classification.

Our parametrisation is closely related to the M2 model proposed in ~\citep{kingma+14}, but has advantages in reduction of neural computations required in learning and inference and also in interpretability of the generative properties. We will show that our parametrisation of the variational autoencoder is a simple, yet an effective one.

\iffalse
Our parametrisation differs from the M2 model proposed in~\citep{kingma+welling14,rezende+14}, by removing the class label from the input to the recognition and generative networks, and instead introducing a class speficic bias to the latent mean parameter. This allows for a simple, yet effective, method of modelling the class conditionality of latent features. This method also reduces the number of neural computations required in unsupervised and semi-supervised learning, as marginalization over class labels requires only one pass over the recognition network.
\fi

The structure of the remaining paper is as follows: The description of the models and methods 
developed, and related work are described in the following section (2). The experiments conducted and the results obtained are described then in section 3. Finally, section 4 summarizes our main results and outlines plans for future work.

\section{Models and Methods}

\subsection{Proposed variational auto-encoder}
We consider a setting, where we have a set of observed data $\mathbf{X}=\{\mathbf{x}^{(1)},\mathbf{x}^{(2)},...,\mathbf{x}^{(n)}\}$ and each observation can be associated with one of $L$ different labels. Elements of $\mathbf{X}$ are usually vectors, such that $\mathbf{x} \in \mathbb{R}^{D}$ where $D$ is the dimensionality of $\mathbf{x}$. The class of an observation $\mathbf{x}^{(i)}$ is indicated by a label $y^{(i)}\in\{0,1,...L-1\}$. These labels are not necessarily given for the whole set of observations and often there is more unlabeled than labeled data. Given an associated label $y^{(i)}$, an observation $\mathbf{x}^{(i)}$ is generated by a latent variable $\mathbf{z}^{(i)}\in\mathbb{R}^{K}$, where $K$ is the dimensionality of the latent variable.

\iffalse
We propose a probabilistic generative model, which considers each observation $\mathbf{x}^{(i)}$ to be generated by a class conditionally distributed latent variable $\mathbf{z}^{(i)}\in\mathbb{R}^{K}$, where $K$ is the dimensionality of the latent variable. A graphical illustration of the model is shown in Figure~\ref{fig:semi-supervised-graphical}.
\fi

The joint probability distribution of a data example $\mathbf{x}$, associated latent variables $\mathbf{z}$ and unobserved class variable $y$, factorizes under our generative 
model as follows:
\begin{equation*}\label{eq:generativeJoint}
p_{\theta}(\mathbf{x},y,\mathbf{z})=p_{\theta}(y)\condProbPP{\mathbf{z}}{y}\condProbPP{\mathbf{x}}{\mathbf{z}},
\end{equation*}
where we assume specific forms for the distributions, and the parameters of the distribution $\condProbPP{\mathbf{x}}{\mathbf{z}}$ are defined by a feed-forward neural 
network mapping; a set of examples in a dataset are assumed to be generated i.i.d from it, i.e. $p_{\theta}(\mathbf{X},y,\mathbf{Z})=\prod_{n=1}^{N}p_{\theta}(\mathbf{x}^{(n)},y^{(n)},\mathbf{z}^{(n)})$. The associated recognition model defines the distribution of the joint configuration of latent variable units and the class variable given a data example as follows: 
\begin{equation*}
\condProbQQ{y,\mathbf{z}}{\mathbf{x}}=\condProbQQ{y}{\mathbf{x}}\condProbQQ{\mathbf{z}}{\mathbf{x},y}
\end{equation*}
where we similarly assume parametrized forms of the distributions, and the parameters for both of the distributions are defined by their specific feed-forward neural network mappings. These parameters are contained in $\{\theta,\phi\}$ for the generative - and recognition models respectively.

The forms of these distributions are defined as:
\begin{eqnarray*}
p_{\theta}(y)&=&\operatorname{Categorical}(y)\\
\condProbPP{\mathbf{z}}{y}&=&\operatorname{Normal}(\mathbf{z};{\boldsymbol{\mu}_y},\operatorname{Diag}(\{\boldsymbol{\sigma}_{y}^{2}\}))\\
\condProbQQ{\mathbf{z}}{\mathbf{x},y}&=&\operatorname{Normal}(\mathbf{z};f_{\phi}(\mathbf{x})+\mathbf{b}_{y},\operatorname{Diag}(g_{\phi}(\mathbf{x}))),
\end{eqnarray*}
where $\boldsymbol{\mu}_y$, and $\boldsymbol{\sigma}_{y}^{2}$ define the class-conditional mean and variance vectors for generative model latent variable prior, and 
$f_{\phi}(\mathbf{x})+\mathbf{b}_{y}$ and $g_{\phi}(\mathbf{x})$ those for the distribution defined by the recognition network.

The functional form of the generative model distribution is chosen according to the application. In our experiments we use binary and continous data, and thus there are two different generative distributions we use, Bernoulli for binary data, and Normal for continuous data. The distribution for the binary data is defined as follows:
\begin{equation*}
\condProbPP{\mathbf{x}}{\mathbf{z}}=\operatorname{Bernoulli}(\mathbf{x};\boldsymbol{\mu}_{\theta}(\mathbf{z})),
\end{equation*}
where $\boldsymbol{\mu}_{\theta}(\mathbf{z})$ defines the success/activation probability of the distribution. The distribution for the continuous data is on the other hand defined as follows:
\begin{equation*}
\condProbPP{\mathbf{x}}{\mathbf{z}}=\operatorname{Normal}(\mathbf{x};F_{\theta}(\mathbf{z}),\operatorname{Diag}(G_{\theta}(\mathbf{z}))),
\end{equation*}
where the mean and variance are mapped similarly as in the recognition network, using a function defined by a neural network.

\begin{figure}
\vskip 0.2in
\begin{center}
\includegraphics[width=0.9\columnwidth]{\figureDir/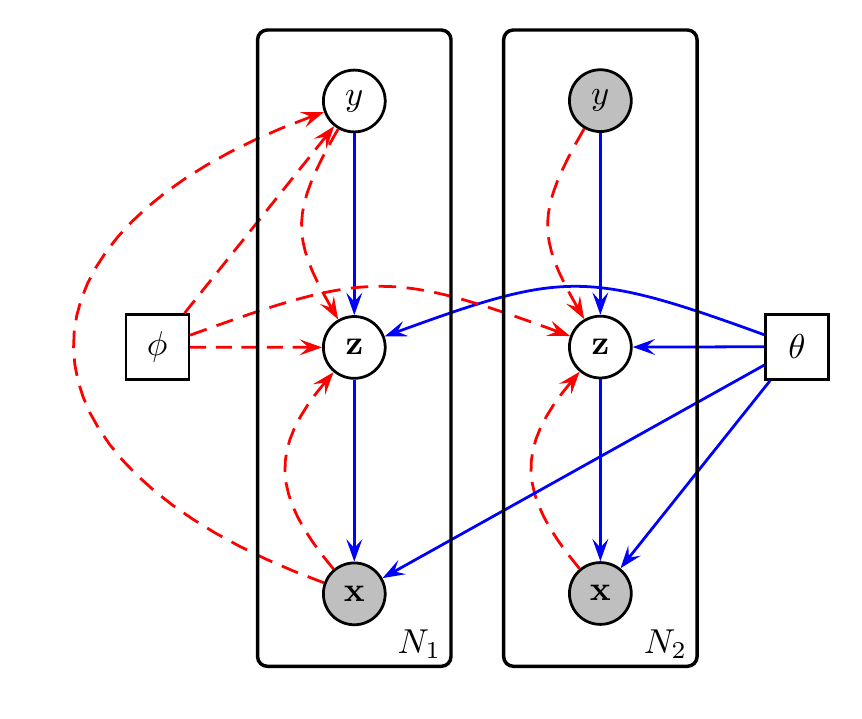}
\caption{Our generative model using the plate notation. The numbers $N_{1}$ and $N_{2}$ denote the task-dependent number of unlabeled and labeled examples, either of which may be zero. Blue lines represent the generative network pathways and dashed red lines represent the recognition network pathways.}\label{fig:semi-supervised-graphical}
\end{center}
\vskip -0.2in
\end{figure}

As in a standard VAE~\citep{kingma+welling14}, the learning happens by optimizing a stochastic estimate of the (standard) evidence lower bound. The bound under the model with observed labels becomes
\begin{small}
\begin{equation}
\mathcal{L}=\mathbb{E}_{\condProbQQ{\mathbf{z}}{\mathbf{x},y}}[\log \condProbPP{\mathbf{x}}{\mathbf{z}}]-\operatorname{{D}_{KL}}[\condProbQQ{\mathbf{z}}{\mathbf{x},y} || \condProbPP{\mathbf{z}}{y}],
\end{equation}
\end{small}
where $\operatorname{{D}_{KL}}$ denotes the KL-divergence, and under the model with unobserved labels, the evidence lower-bound becomes
\begin{small}
\begin{equation}
\begin{split}
\mathcal{U}&=\mathbb{E}_{\condProbQQ{y}{\mathbf{x}}}[\mathbb{E}_{\condProbQQ{\mathbf{z}}{\mathbf{x},y}}[\log \condProbPP{\mathbf{x}}{\mathbf{z}}]-\operatorname{{D}_{KL}}[\condProbQQ{\mathbf{z}}{\mathbf{x},y} || \condProbPP{\mathbf{z}}{y}]]\\
&=\sum_{y}\condProbQQ{y}{\mathbf{x}}\mathbb{E}_{\condProbQQ{\mathbf{z}}{\mathbf{x},y}}[\log \condProbPP{\mathbf{x}}{\mathbf{z}}] + \mathcal{H}(q_{\phi}(y|\mathbf{x}))\\
&~~~-\sum_{y}\condProbQQ{y}{\mathbf{x}}\operatorname{{D}_{KL}}[\condProbQQ{\mathbf{z}}{\mathbf{x},y} || \condProbPP{\mathbf{z}}{y}].
\end{split}
\end{equation}
\end{small}
\vspace{-1mm}
However, in our experiments we considered a different (and looser) lowerbound than $\mathcal{U}$, denoted by ~$\mathcal{U}^{\ast}$, defined as ~$\mathcal{U}$ without the entropy term $\mathcal{H}(q_{\phi}(y|\mathbf{x}))$.\\

%Missing entropy term from U: + \mathcal{H}(q_{\phi}(y|\mathbf{x}))
\CUT{
\begin{small}
\begin{equation}
\begin{split}
\mathcal{L}&=\mathbf{E}_{\condProbQQ{\mathbf{z}}{\mathbf{x},y}}[log \condProbPP{\mathbf{x}}{\mathbf{z}}]\\&-KL[\condProbQQ{\mathbf{z}}{\mathbf{x},y} || \condProbPP{\mathbf{z}}{y}]
\\\\\mathcal{U}&=\mathbf{E}_{\condProbQQ{y}{\mathbf{x}}}[\mathbf{E}_{\condProbQQ{\mathbf{z}}{\mathbf{x},y}}[log \condProbPP{\mathbf{x}}{\mathbf{z}}]\\&-KL[\condProbQQ{\mathbf{z}}{\mathbf{x},y} || \condProbPP{\mathbf{z}}{y}]]
\\&=\sum_{y}\condProbQQ{y}{\mathbf{x}}(\mathbf{E}_{\condProbQQ{\mathbf{z}}{\mathbf{x},y}}[log \condProbPP{\mathbf{x}}{\mathbf{z}}]\\&-KL[\condProbQQ{\mathbf{z}}{\mathbf{x},y} || \condProbPP{\mathbf{z}}{y}])
\end{split}
\end{equation}
\end{small}

\begin{flalign*}
\nonumber \mathcal{L} &= \mathbf{E}_{\condProbQQ{\mathbf{z}}{\mathbf{x},y}}[log \condProbPP{\mathbf{x}}{\mathbf{z}}] - KL[\condProbQQ{\mathbf{z}}{\mathbf{x},y} || \condProbPP{\mathbf{z}}{y}]\\ \\
\nonumber \mathcal{U}&=\mathbf{E}_{\condProbQQ{y}{\mathbf{x}}}[\mathbf{E}_{\condProbQQ{\mathbf{z}}{\mathbf{x},y}}[log \condProbPP{\mathbf{x}}{\mathbf{z}}]  \\
\nonumber -KL[\condProbQQ{\mathbf{z}}{\mathbf{x},y} || \condProbPP{\mathbf{z}}{y}]]\\
\nonumber &=\sum_{y}\condProbQQ{y}{\mathbf{x}}(\mathbf{E}_{\condProbQQ{\mathbf{z}}{\mathbf{x},y}}[log \condProbPP{\mathbf{x}}{\mathbf{z}}]\\
-KL[\condProbQQ{\mathbf{z}}{\mathbf{x},y} || \condProbPP{\mathbf{z}}{y}])
\end{flalign*}
}
Combining the above results in our semi-supervised objective:
$\mathcal{S}=\mathcal{L}+\mathcal{U}^{\ast}$
, but in practice we also use an additional term, similar to the one used in~\citet{kingma+14}, to train the classifier: 
\begin{equation}
 \mathcal{C}=\mathbb{E}_{p_{\text{emp.}}(\mathbf{x},y)}[\log(\condProbQQ{y}{\mathbf{x}})],
\end{equation}
where $p_{\text{emp.}}(\mathbf{x},y)$ denotes the empirical distribution over labeled data. We introduce this term, because in the original objective only the unlabeled objective is dependent on $\condProbQQ{y}{\mathbf{x}}$. This additional term trains the classifier with the labeled samples used in the supervised lower bound calculation.

We add the additional term with a multiplier to the semi-supervised objective to get the full objective:
\begin{equation}
\mathcal{S}=\mathcal{L}+\mathcal{U}^{\ast}+\alpha\cdot\mathcal{C}
\end{equation}
As usual with VAEs, we cannot compute the lower bounds exactly, but take Monte Carlo-estimates of them, and by the reparametrization trick~\citep{kingma+welling14,rezende+14}, we are able to optimize the estimates so that the generation and inference networks parameters are tuned concurrently, i.e. in concerted fashion. 
\subsection{Nominal classification}
In a nominal classification task, we try to predict a discrete class label $y^{(i)}\in\{0,1,...L-1\}$ for an observation $\mathbf{x}^{(i)}$. We can infer probabilities for each class label, given an example, using a nominal classifier. Usual choice for encoding the categorical distribution under a neural network is via the softmax:
\begin{equation}
\operatorname{softmax}(y_{k};\mathbf{x})=\frac{\exp{\{\transpose{\mathbf{W_{k}}}\mathbf{h}{(\mathbf{x})}+b_{k}\}}}{\sum_{\ell}\exp{\{\transpose{\mathbf{W}_{\ell}}\mathbf{h}{(\mathbf{x})}+b_{\ell}\}}},
\end{equation}
where $\mathbf{W_{k}}$ and $b_{k}$ denote the class specific mapping vector and bias and the $\mathbf{h(\mathbf{x})}$ denote a neural network mapping of $\mathbf{x}$. We chose softmax to encode the categorical distribution of $q_{\phi}(y|\mathbf{x})$. The classification of an example $\mathbf{x}^{(i)}$ is then by evaluating the probabilibities with it, and the class is set to the label
which gives the highest probability under it, i.e.
\begin{equation}
\nonumber y^{(i)}_{\ast}=\argmax_{k} q_{\phi}(y^{(i)}=k|\mathbf{x}^{(i)}).
\end{equation}
\subsection{Ordinal classification}
Ordinal classification, also called ordinal regression, is a classification task where the discrete class labels have an order amongst them. In such setting, the classification loss is not only a function of the misclassification, but also the distance from the misclassified label to the correct one~\citep{gutierrez+16}. Example of such a case could be diagnosis of a disease, where classifying the grade of disease lower than the actual grade, would cause progressively higher risk of fatality, the further the predicted grade is from the correct one. On the other hand, classifying the grade higher than the actual grade would cause progressively higher costs resulting from unnecessary treatment.~\citep{pedregosa+17}

Multiple different techniques exist in the domain of ordinal classification. Using nominal classification methods is a simple approach and often ignores any ordinal features present in the data. For example, standard softmax classifier tries to partition the features in input data to orthogonal directions. Only the features in the direction of a class-dependent softmax weight vector will cause a response. This method clearly fails in a situation where the features are all in the same direction, but the magnitude separates different classes.

Choices for ordinal classifiers include (but is not restricted to): standard regression with prediction determined by nearest label to the output~\citep{pedregosa+17}~\citep{gutierrez+16}, using the expected value of nominal classifier prediction and minimizing the squared distance to the target~\citep{beckham+pal16}, cumulative link models that use scalar mappings to estimate cumulative probabilities for class labels, such as $\condProbP{y \preceq \mathbf{y_{k}}}{\mathbf{x}} = f(\mu_{k}-W^{T}x)$~\citep{gutierrez+16}, Gaussian Processes (GP)~\citep{chu+ghahramani05} and Support Vector Machines (SVM)~\citep{herbrich+99}.

\iffalse
Standard regression can be used in ordinal classification by treating the classification task as a regression problem. One such solution is to use least squares algorithm for the training process and for predictions round the function output to nearest label value~\citep{pedregosa+17}~\citep{gutierrez+16}. One similar approach is to use standard nominal classifier to calculate individual class probabilities and then calculate the expected value of the label. Objective is to minimize the squared difference between this expected value of the label and the true label~\citep{beckham+pal16}. Other techiniques, like cumulative link models, use scalar mappings to estimate cumulative probabilities for class labels. Such a model could be defined as
\\ $\condProbP{y \preceq \mathbf{y_{k}}}{\mathbf{x}} = f(\mu_{k}-W^{T}x)$
,where the function $f$ could be chosen as sigmoid-function for example. Also Gaussian Processes (GP)~\citep{chu+ghahramani05} and Support Vector Machines (SVM)~\citep{herbrich+99} have been used in ordinal regression.
\fi

What many ordinal methods have in common is the use of a scalar output function and a set of thresholds, which function as decision boundaries. A set of such thresholds could be, using similar notation as in ~\citet{gutierrez+16}, $\mathbf{b} = (b_{1},b_{2},...,b_{L-1})$.
This set of thresholds defines intervals, usually with the beginning point of first interval set to negative infinity and the end of last interval to infinity. Every interval is associated to predicting a certain class. Using a function with a scalar output, we would predict class $i$ if $f(\mathbf{x}) \in [b_{i-1},b_{i}]$.

For the purpose of enabling marginalization in ordinal domain, we developed an ordinal version of softmax with connection to the forementioned threshold methods. The form of this classifier is as follows: 
\begin{equation}
\operatorname{ordmax}(y=k;\mathbf{x})=\frac{\exp{\{-(\transpose{\mathbf{W}}\mathbf{h(\mathbf{x})}+b-\mu_{k})^{2}/s\}}}{\sum_{\ell}\exp{\{-(\transpose{\mathbf{W}}\mathbf{h(\mathbf{x})}+b-\mu_{\ell})^{2}/s\}}},
\end{equation}
where $\mu_{k}<\mu_{k+1}~\forall~k\in[1,2,\ldots,K-1]$. Similarly as softmax, the partition term is the sum of all the individual terms in this classifier, and the assigned probability to one class label is the weight the corresponding term has in the sum. This ensures that the probabilities sum up to one. The magnitudes of individual terms are defined by the learnable mapping $\transpose{\mathbf{W}}\mathbf{h(\mathbf{x})}+b$ which outputs a scalar. The ordinality is introduced to the model, by defining the distribution centres $\mu_k$ as constant scalars with magnitudes increasing with the index. Thus the resulting probabilities have a peak on the class with closest centre to the mapping and the other probabilities decrease progressively the further we get away from this closest class. In our experiments, each the $\mu$'s was set to the numerical value of the corresponding label, shifted so that the mean value of $\mu$'s was zero. 
 
We also introduced the learnable scaling factor $s$ to help in stabilization of the learning process. The point of $s$ is to flatten the initial distribution such that the initial probabilities would be more evenly distributed. As the model becomes more confident in its predictions it can decrease $s$ to concentrate more probability into its predictions. Without $s$ and centering the $\mu$'s, we ran into numerical issues in calculating cross-entropy, as often the log term was rounded into negative infinity.

\subsection{Related work}
The natural comparison of this model is with the generative semi-supervised model M2 described in~\citet{kingma+14}. Their M2 model considers the generative process to be a function of statistically independent $y$ and $\mathbf{z}$ such that $p_{\theta}(y,\mathbf{z}) = p_{\theta}(y)p_{\theta}(\mathbf{z})$, with $p_{\theta}(y)$ categorical – and $p_{\theta}(\mathbf{z})$ standard normal distribution. In our model the latent variables and the class information are not independent, but $\mathbf{z}$ has a conditional normal distribution given $y$. In the M2 model the prior $p_{\theta}(\mathbf{z})$ is a spherical Gaussian, with no learnable parameters, whereas our model $\condProbPP{\mathbf{z}}{y}$ consists of learnable mean vectors and diagonal covariance matrices. 
%In practice we chose a vector of variances in place of a covariance matrix. Since the variances of the latent priori distributions were learnable, we observed that the shapes of these distributions were rarely spherical.

The computational complexity of M2 model, described in~\citet{kingma+14} section 3.3, is stated to be that of $LC_{M1}$, where $C_{M1}$ is the computational complexity of their other model M1 and $L$ is the number of classes. The $C_{M1}$ includes the computational costs of both recognition and generative networks. In contrast to M2 model, our model would have the computational complexity (assuming same network configuration as in their M2 experiments and that summation is negligible compared to neural computations) $\frac{C_{M1}}{2}(L+1)$, as our model requires only one pass over recognition network and the marginalisation is performed by $L$ sums and passes over the generative network.
\iffalse
The M2 model also includes the class label as an input into both the recognition and generative networks. This means that the marginalization over class labels requires $N_{classes}$ computations over the whole network architecture, where $N_{classes}$ is the number of classes. We instead introduce the label information as a bias to the mean parameter of the latent distribution. This results in marginalization being less demanding of neural computations, as our model requires only one pass over the recognition network and then $N_{classes}$ sums and propagations over the generative network. The class specific bias also allows for intuitive implementation of a convolutional variant, where $\mathbf{x}$ and/or $\mathbf{z}$ are feature maps, as the class specific bias can be of arbitarily chosen shape, whereas concatenating the label to the input might be challenging.
\fi

 One another variational autoencoder approach that enables semi-supervised learning is described in~\citet{maaloe+16}, but in addition to the approaches in~\citet{kingma+14} it has not been applied to ordinal classification. Also in contrast to our approach, it defines stochastically deep models, which we have excluded from the scope of this paper.

A similar appearance to our lower bound formulation appears in~\citet{sohn+15}, but in contrast to their CVAE model, ours has no recognition network parameters for the latent prior distribution, which is defined by class conditional parameters and ours has no generative network parameters for the class label $y$. Judging from their lower bound formulation (equation (4)) and their graphical visualisation of conditional dependence (Figure 1), our lower bound is different from theirs.

There has been some previous work on semi-supervised ordinal regression. Supervised Gaussian processes for ordinal regression were used in~\citet{chu+ghahramani05}, and a semi-supervised version was introduced in~\citet{srijith+13}. Also variational Gaussian process auto-encoders were developed in~\citet{stefanos+16} for ordinal prediction of facial action units. Other techniques include kernel discriminant analysis with an additional order forcing term~\citep{perez-ortiz+16}.
\section{Experiments and Results}
In this section we will present some of our results using benchmark datasets on both nominal - and ordinal classification. We will also present some visualizations to better explain the inner workings of the model.
We used the Python-based Theano-framework in our algorithm implementations~\citep{theano-short}.
\subsection{Modelling and classifying MNIST-digit data}
The MNIST dataset consists of 28 x 28 grayscale images of handwritten digits. We obtained the dataset directly from Yann LeCun's website~\citep{mnist-database-website} and divided the given training set onto a training and a validation set of size 50 000 and 10 000, respectively.  We also used the data preprocessing scheme detailed in the~\citet{kingma+14}. This preprocessing included data normalization to interval [0,1] and sampling individual pixels from a Bernoulli distribution using these normalized pixel intensities as success probabilities.

\subsubsection{Generative modelling}
We experimented on generative modelling with multiple different settings of hidden layers, hidden neurons, latent neurons and fractions of labelled data. Common settings for all our trained models were the use of softplus non-linearity and sigmoid function in the generative model output. Sigmoid function was used, because generative model distribution was chosen to be Bernoulli. This way the output neurons can be always interpreted as individual Bernoulli success probabilities and standard cross-entropy used in the objective function. In optimization, we used the Adam algorithm described in~\citet{kingma+ba15}. 

\begin{figure*}
\vskip 0.2in
\begin{center}
\includegraphics[width=0.9\textwidth]{\figureDir/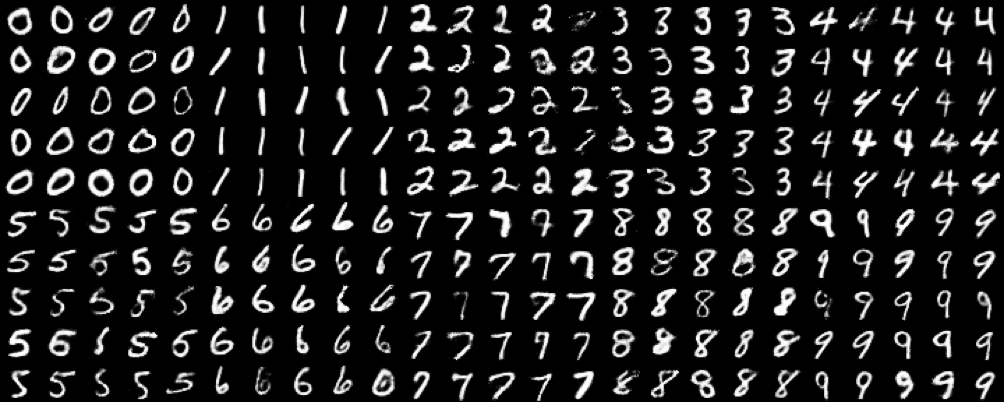}
\caption{Sampling from an MNIST-model: Images of the sampling distribution  $\condProbPP{\mathbf{x}}{\dot{\mathbf{z}}}$ mean, given 25 random samples $\dot{\mathbf{z}}$ from $\condProbPP{\mathbf{z}}{y}$, the distribution of the latent variables conditional on the class label $y$, under each of the 10 classes.\label{fig:mnist-model-regular-synthesis}}
\end{center}
\vskip -0.2in
\end{figure*}
We observed that the visual quality of the samples was enhanced with adding layers to the model and also by increasing the amount of labelled data available to the model. The resulting digits, shown in Figure~\ref{fig:mnist-model-regular-synthesis}, represent samples from the learned latent prior distributions of a model with two hidden layers on all networks and 10000 labelled samples available to the model.

\begin{figure*}
\vskip 0.2in
\begin{center}
\includegraphics[width=0.9\textwidth]{\figureDir/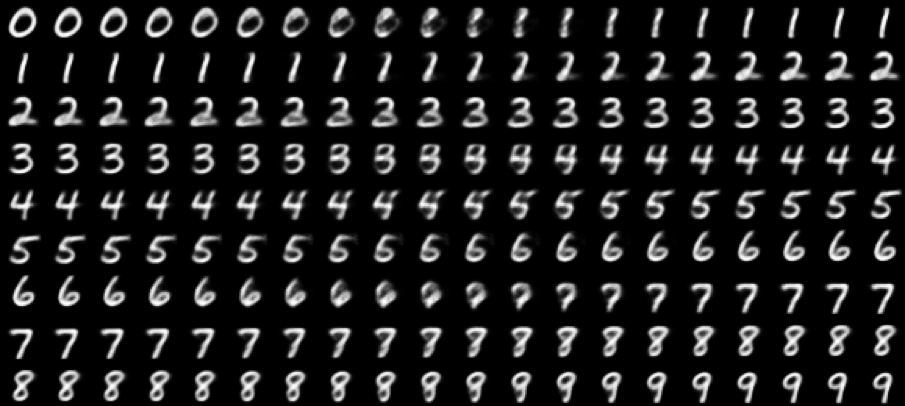}
\caption{Class-interpolated sampling from an MNIST-model: Images of the sampling distribution $\condProbPP{\mathbf{x}}{\mathbf{z}}$ mean under multiple configurations of (50-dimensional) $\mathbf{z}$, with one configuration per image. The configurations per each row start (first column) from the mean of $\condProbP{\mathbf{z}}{y=k}$ (the probability distribution of the latent variables configuration conditional on the class label $y$ being $k$), and linearly interpolate to (last column) the mean of $\condProbPP{\mathbf{z}}{y=k+1}$ (the probability distribution of the latent variables configuration conditional on the class label $y$ being $k+1$). We can see that the interpolation results in semantically smooth digit morphs.\label{fig:mnist-model-latent-interpolation-synthesis}}
\end{center}
\vskip -0.2in
\end{figure*}

We observed that even thought removing layers and the amount of labelled data degraded the visual quality of the samples produced by the model, the reconstructions of interpolations between the latent distribution centres were well defined. We can see from Figure~\ref{fig:mnist-model-latent-interpolation-synthesis}, how the linear interpolations in 50 dimensional latent space result in transformations from the original digit (shown in the left) towards the target digit (shown in the right). This model had one hidden layer on all networks with 500 hidden neurons and 1000 labelled examples available. The effect of latent space interpolation has been also considered in~\citet[see Fig. 11]{nash+williams17} but under different semantics.

\begin{figure*}
\vskip 0.2in
\begin{center}
\includegraphics[width=0.9\textwidth]{\figureDir/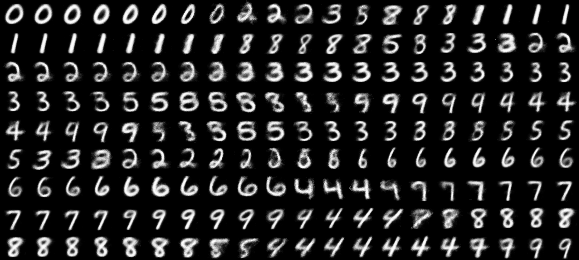}
\caption{Class-interpolated sampling from model with 2-dimensional latent space. Visual inspection of Figure 4 explains the different digits appearing in the interpolation, as we pass through other distributions in latent space.\label{fig:mnist-model-latent-interpolation-synthesis-2D}}
\end{center}
\vskip -0.2in
\end{figure*}

\begin{figure}
\vskip 0.2in
\begin{center}
\includegraphics[width=0.9\columnwidth]{\figureDir/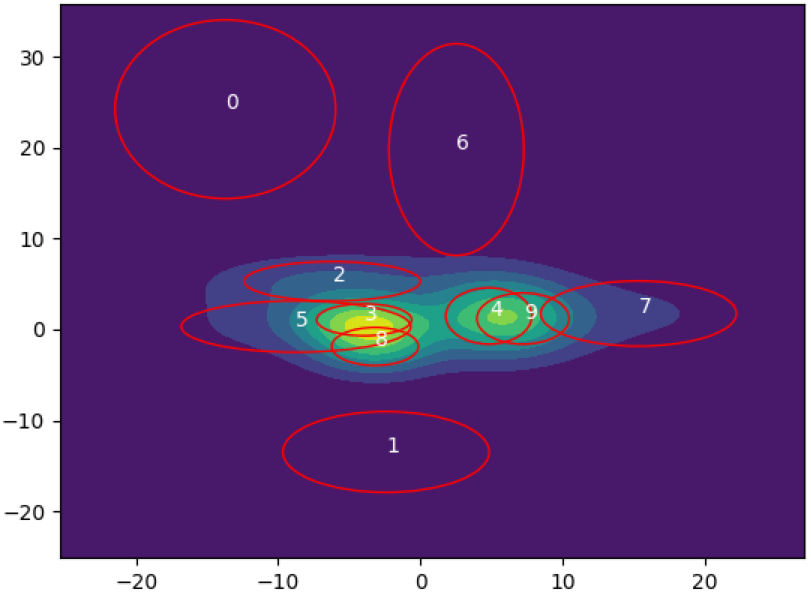}
\caption{Class-conditional 2D-latent variable prior-distributions, and their mixture. Each (red) ellipse marks equiprobable points under a conditional distribution, with distribution mean at the center (and respective class label nearby), and points on the minor and major axis a standard deviation away from it. The background colouring encodes the mixture density.\label{fig:mnist-model-2D-latent-priors}}
\end{center}
\vskip -0.2in
\end{figure}

Decreasing the dimensionality of $\mathbf{z}$ to 2 allows us to visualize the distributions in the latent space. We observed that in this case, keeping the last network configuration resulted in poor performance, but by compensating the loss of latent dimensionality by adding one additional hidden layer, more hidden neurons and increasing the number of labelled samples increased the model performance to comparable results with the previous model. However the number of labelled examples had to be increased up to 40000. We can see from Figure~\ref{fig:mnist-model-2D-latent-priors} that digits with similar features seem to get close to one another. Interpolation between consecutive digits in this lower dimensional space shows how the latent distributions get more interleaved, as different digits appear on the way from one digit to another in Figure~\ref{fig:mnist-model-latent-interpolation-synthesis-2D}.

\subsubsection{Nominal classification}
In semi-supervised classification task, we chose our network complexity to match that of M2 in~\citet{kingma+14}. This means one hidden layer in both recognition and generative networks, both with 500 neurons. The latent $\mathbf{z}$ consists of 50 neurons and the classifier network is also one hidden layer with 500 neurons and a softmax classifier. As in generative modelling, the activations were chosen as softplus activations aside from the sigmoid activation on the generative network output, both similar choices to ~\citet{kingma+14}. We chose to only compare our model to the M2 as both have similar stochastic depth, only one layer of stochastic variables, and by the same configuration they also have similar number of learned parameters.

The results of classification errors on different number of labelled data can be seen in Table~\ref{tab:mnist-classification}. We can see that our model achieves lower error rates on average in the cases of 100, 600 and 3000 labelled examples, however ours has higher standard deviation on the cases 100 and 600. We can also see that there is consistency in our model performance, as the error always decreases on average by adding more labelled data to the supervised set. 

\iffalse
We compared our model to other semi-supervised classification methods on the MNIST dataset. We used similar model complexity as the M2-model in~\citet{kingma+14}. This means we had one hidden layer on both the recognition - and the generative model, both with 500 neurons and 50-dimensional latent variable. The classifier we used was a standard softmax-classifier with its own network that had one 500-neuron hidden layer. In our experiments we used a similar additional objective for the classifier as in~\citet{kingma+14}. This additional term tries to minimize the classification loss of the labeled examples. This way the classifier can be directed to have certain softmax weights for predetermined class and thus the overall classification loss can be evaluated without further analysis of the classifier. In our experiments the classification loss was very dependent on this additional term. With using this term as is, the classifier associated the different classes in a random manner to different softmax weights. Introducing a multiplier $\alpha$ with a high value, corrected this problem. 
\fi

\begin{table}
{\footnotesize
\caption{Semi-supervised MNIST classification statistics: Error using different amounts of labelled examples with the training set, comparing the proposed approach and results of the M2-model in~\citet[reproduced from Table 1]{kingma+14}. MEAN denotes the mean (lowest highlighted), and STD the standard deviation, of classification error. We have used 10 different runs (each choosing labeled examples randomly) to compute our results.\label{tab:mnist-classification}}
\vskip 0.15in
\centering
\begin{tabular}{llll}
\toprule
& \multicolumn{2}{c}{Result statistics} & Amount labeled\\
\cmidrule(r){2-3}
VAE  & {\bf MEAN} & STD &  \\
\midrule
 & &  & 100 \\
M2 & 11.97 & 1.71 & \\
Us & \textcolor{blue}{\bf 8.16} & 3.22  &  \\
\midrule
 & &  & 600 \\
M2 & 4.94 & 0.13 & \\
Us & \textcolor{blue}{\bf 4.62} & 0.23  &  \\
\midrule
 & &  &1000 \\
M2 & \textcolor{blue}{\bf 3.60} & 0.56 & \\
Us & 4.32 & 0.22  &  \\
\midrule
 & &  & 3000 \\
M2 & 3.92 & 0.63  & \\
Us & \textcolor{blue}{\bf 3.85} & 0.25  &  \\
\bottomrule
\end{tabular}
}
\vskip -0.1in
\end{table}

\subsection{Modelling and classifying Stocks-data}
We tested the flexibility of our model by implementing it to semi-supervised ordinal regression. We used the ordinal regression benchmark dataset 'stock domain'~\citep{stocks-database-website}. This dataset consists of 950 samples, with 9 features and an ordinal label. According to the authors~\citet{chu+ghahramani05}, this dataset was created from a standard regression problem, by binning the target regression interval into 5 equal length bins each representing one label. We divided these 950 samples 10 times into partitions of 600 training samples and 350 test samples. We used these partitions to train and evaluate model performance with multiple different fractions of labeled training data. In our experiments normalizing the data to zero-mean and unit variance proved to have a stabilizing effect on the learning process.

\subsubsection{Generative modelling}
The features of this dataset are all continuous values, thus the natural choice for the generative network output distribution is Gaussian. We also used a data-independent global variance for the Gaussian distribution, instead of a neural network mapping from the latent variables. This variance parameter was also learned by the model. Since there are only 9 features in the data, we used considerably smaller network than what we used in the MNIST experiments. The network had 1 hidden layer on both recognition – and generative model and both had 7 neurons. We used only one neuron for the latent feature layer and for the non-linearity, we chose hyperbolic tangent (tanh) activation.

\subsubsection{Ordinal classification}

The fractions of labeled data was set to similar values as used in~\citet{srijith+13}. The fractions represent the ratio of labeled – to all training data. These fractions were created by sampling a subset of training data for which we kept the labels and discarded them for the rest. For comparison we also trained the model on the ordinal data using standard softmax with the rest of the model kept the same. Both classifiers also had a one hidden layer MLP connection from the data with hyperbolic tangent nonlinearity.

\begin{table}
{\footnotesize
\caption{Stock domain-dataset classification statistics: Error using different fractions of labelled to all data, comparing the two approaches. MEAN denotes the mean (lowest highlighted), and STD the standard deviation, of classification (zero-one) error.\label{tab:stocks-classification}}
\vskip 0.15in
\centering
\begin{tabular}{llll}
\toprule
& \multicolumn{2}{c}{Result statistics} & Fraction (labeled)\\
\cmidrule(r){2-3}
Softmax  & {\bf MEAN} & STD &  \\
\midrule
 & &  &0.05 \\
Standard & 0.354 & 0.069 & \\
Ordinal & \textcolor{blue}{\bf 0.333} & 0.063  &  \\
\midrule
 & &  &0.10 \\
Standard & 0.283 & 0.044 & \\
Ordinal & \textcolor{blue}{\bf 0.258} & 0.038  &  \\
\midrule
 & &  &0.15 \\
Standard & 0.258 & 0.041 & \\
Ordinal & \textcolor{blue}{\bf 0.219} & 0.037  &  \\
\midrule
 & &  &0.20 \\
Standard & 0.224 & 0.041  & \\
Ordinal & \textcolor{blue}{\bf 0.206} & 0.030  &  \\
\midrule
 & &  & 0.25 \\
Standard & 0.217 & 0.029  & \\
Ordinal & \textcolor{blue}{\bf 0.198} & 0.026  &  \\
\midrule
 & &  & 0.30 \\
Standard & 0.203 & 0.025 & \\
Ordinal & \textcolor{blue}{\bf 0.189} & 0.017  &  \\
\bottomrule
\end{tabular}
}
\vskip -0.1in
\end{table}

\begin{figure}
\vskip 0.2in
\begin{center}
\includegraphics[width=0.9\columnwidth]{\figureDir/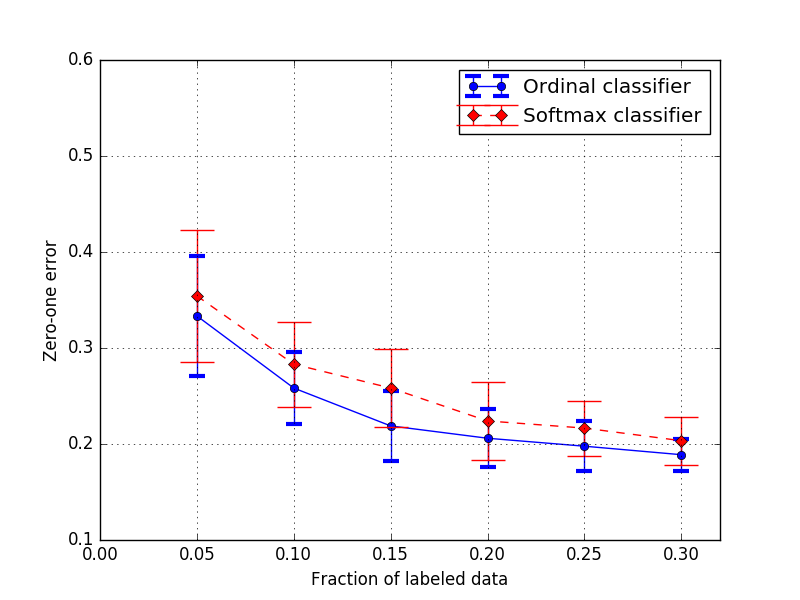}
\caption{Stock domain-classification statistics: Classifier accuracy using different fractions of labelled to all data, comparing the two approaches.\label{fig:stocks-classification}}
\end{center}
\vskip -0.2in
\end{figure}

We present the test set errors for both models in Table~\ref{tab:stocks-classification}. We can see that in all cases the ordinal classifier had better accuracy than the softmax classifier using the same amount of labeled data. However, in most cases the difference is within the margin of one standard error. Reader is also recommended to compare our results shown in Figure~\ref{fig:stocks-classification} to the ones provided in~\citet{srijith+13} - Figure 2. By visual inspection, we can see similar performance in our model.

In the Fig.~\ref{fig:ordinal_latent_space1D} and~\ref{fig:softmax_latent_space1D} we can see examples of the one dimensional learned priori distributions for this ordinal data. For the ordinal classifier case there seems to be high mixing of the classes 3 and 4. In the softmax classifier case, we see that classes 2 and 3 have very flat distributions and the peaks seem to match the peaks of classes 1 and 5. Class one has very concentrated mass which results in a peak in the figure.

\CUT{
\begin{figure}
\vskip 0.2in
\begin{center}
\includegraphics[width=0.9\columnwidth]{\figureDir/stocks-classification-ordinal-softmax.png}
\caption{Stock domain-classification statistics: Classifier accuracy using different fractions of labelled to all data, {\bf ordinal-softmax} unit-based approach.\label{fig:stocks-classification-ordinal-softmax}}
\vskip -0.2in
\end{figure}

\begin{figure}
\vskip 0.2in
\begin{center}
\includegraphics[width=0.9\columnwidth]{\figureDir/stocks-classification-standard-softmax.png}
\caption{Stock domain-classification statistics: Classifier accuracy using different fractions of labelled to all data, {\bf standard softmax} unit-based approach.\label{fig:stocks-classification-standard-softmax}}
\end{center}
\vskip -0.2in
\end{figure}
}

\begin{figure}
\vskip 0.2in
\begin{center}
\includegraphics[width=0.9\columnwidth]{\figureDir/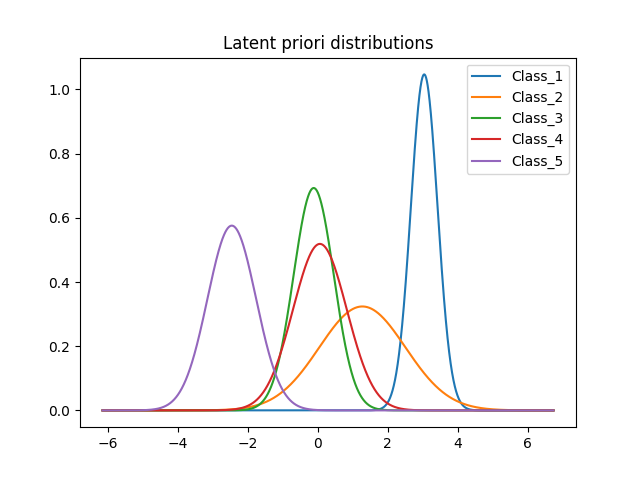}
\caption{One dimensional latent space, Ordinal classifier: Ordering of the latent space using ordinal classifier.\label{fig:ordinal_latent_space1D}}
\end{center}
\vskip -0.2in
\end{figure}

\begin{figure}
\vskip 0.2in
\begin{center}
\includegraphics[width=0.9\columnwidth]{\figureDir/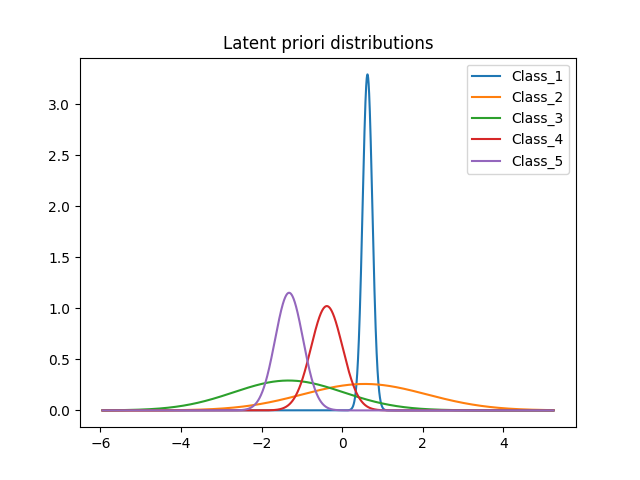}
\caption{One dimensional latent space, Softmax classifier: Ordering of the latent space using softmax classifier.\label{fig:softmax_latent_space1D}}
\end{center}
\vskip -0.2in
\end{figure}

\section{Summary and Discussion}

We have introduced a novel variational autoencoder, which we have shown to be capable of learning features from both nominal - and ordinal data in semi-supervised manner, with results comparable to similar complexity models specifically designed for these types of data. Our model also enables easy interpretation of the learned class-wise features, as each class has their own distinct distributions.

Our main future work is on retinal image analysis, including semi-supervised retinal image classification, for which we have developed a convolutional variant of our model. Because the label information is introduced as a bias to the latent mean parameter, the fully-connected networks can be replaced with convolutional ones and the model parametrisation is still intuitive. One interesting avenue of further research could include a deep stochastic variant of our model, with multiple layers of stochastic variables $\mathbf{z}$. This would also make the model more comparable to the forementioned deep stochastic models. Such extensions could be useful in an another interesting avenue of further research, developing extensions for semi-supervised retinal image (semantic) segmentation.  

\iffalse
 The convolutional form  Our model allows for straightforward implementation of a convolutional variant, as the label information is included as a bias rather than input to the networks. We also excluded deeper stochastic models from the scope of this paper, as our model has currently only one layer of stochastic variables. Thus one interesting avenue could be implementation of multiple layers of stochasticity into our model.

We have introduced a novel variational autoencoder, which we have shown to be capable of learning features from both nominal - and ordinal data in semi-supervised manner, with results comparable to models specifically designed for these types of data. Our model also enables easy interpretation of the learned class-wise features, as each class has their own distinct distributions.

Our main future work direction is on extensions for considering semi-supervised retinal image segmentation (where we might have a label unit per pixel), and also (global) classification. The main architectural extension is then to consider translation equivariant (convolutional) extensions. It is possible that for effective performance stochastically deeper architectures, combined with tighter variational bounds for model optimization, would become more relevant.
\fi
As we saw, e.g.~with the 2D latent space illustration of Fig.~\ref{fig:mnist-model-2D-latent-priors}, in low dimensions the latent distributions may get significantly interleaved. One avenue of future research is to introduce an objective that would separate these distributions in a controlled manner during the training process. One such objective could be by the multi-class version of the Fisher's linear discriminant applied to the latent prior distributions to encourage their separation. This way it could be possible to get good class separation, using lower number of latent neurons.

\section*{Acknowledgements}
The initial idea of having a hierarchical prior specification for a VAE similar to as described in section 2.1. was conceived by JJK in collaboration with Aaron Courville and Yoshua Bengio in early summer - autumn 2014, independent of~\citet{kingma+14}. JJK was then a post-doctoral fellow at LISA (now MILA) of the University of Montreal, and is very thankful for the obtained support.

\bibliography{pubForumStrings,references}
\bibliographystyle{icml2018modified}
\end{document}